\documentclass[11pt]{article}

\def\paperDraft{}

\usepackage[final]
{acl}

\usepackage{times}
\usepackage{latexsym}

\usepackage[T1]{fontenc}

\usepackage[utf8]{inputenc}

\usepackage{microtype}

\usepackage{inconsolata}

\usepackage{graphicx}

\usepackage{custom}
\newcolumntype{C}[1]{>{\centering\arraybackslash}p{#1}}
\RequirePackage{xltabular}

\usepackage{placeins}
\usepackage{stfloats} 

\usepackage[capitalise]{cleveref}

\usepackage{newunicodechar}

\newcommand{\triangleshift}{-.5ex} 

\newcommand{\triUL}{%
  \raisebox{\triangleshift}{%
    \tikz[baseline=-0.55ex,x=1ex,y=1ex]\fill (0,1)--(0,0)--(1,1)--cycle;%
  }%
}

\newcommand{\triLR}{%
  \raisebox{\triangleshift}{%
    \tikz[baseline=-0.55ex,x=1ex,y=1ex]\fill (1,0)--(0,0)--(1,1)--cycle;%
  }%
}

\newunicodechar{◤}{\triUL}
\newunicodechar{◢}{\triLR}

\title{Do we need to answer that question? Salience and Answerability of Potential Questions in Naturalistic Dialogue}

\author{Amandine Decker$^{1,2}$, Maxime Amblard$^{1}$ \and Ellen Breitholtz$^2$\\
        \textsuperscript{1}Université de Lorraine, CNRS, Inria, LORIA, F-54000 Nancy, France \\ \href{mailto:amandine.decker@loria.fr,maxime.amblard@loria.fr}{\texttt{\{amandine.decker, maxime.amblard\}@loria.fr}} \\ \textsuperscript{2}University of Gothenburg, CLASP \\ \href{mailto:ellen.breitholtz@ling.gu.se}{\texttt{ellen.breitholtz@ling.gu.se}}}

\begin{document}
\maketitle
\begin{abstract}
We empirically investigate Question Under Discussion based modelling in naturalistic dialogue by studying whether the salience of generated potential questions predicts their subsequent resolution. Building on \citet{wu2024questions}, we construct a dataset of 7,124 questions automatically generated  from utterances and preceding context from the British National Corpus, and annotated for salience and answerability. We find a robust but low positive correlation between salience and answerability in dialogue, indicating that more salient questions are more likely to be addressed. However, this effect is markedly weaker than in monologic text, suggesting that conversational structure is less predictable. We further observe that structured interactions exhibit stronger alignment between annotators than less organised dialogues. 
\end{abstract}


\section{Introduction}

Discourse structure is a fundamental aspect of 
language use, yet it remains challenging to characterise in a way 
theoretically precise and empirically tractable. One influential approach to modelling discourse structure is the theory of \textit{Questions Under Discussion} (QUDs, \citealp{ginzburg1995resolving,van1995discourse,roberts1996information}), which represents discourse as organised around a dynamically evolving set of questions that conversational participants may 
require answers for 
during a dialogue. On this view, discourse proceeds by addressing questions whose possible answers define the space of relevant contributions at each point 
of interaction.

The QUD framework 
provides a unified 
view on discourse organisation: it can be used retrospectively to reconstruct the structure and content of an interaction, or predictively to model expectations about how discourse will develop. In the predictive setting, so-called \textit{potential questions} (PQs, \citealp{Onea2016PotentialQuestions}) are used to represent candidate issues that may become active QUDs as the discourse unfolds.

Despite its theoretical appeal, operationalising the QUD framework in natural data 
is difficult. The space of relevant questions is not fixed, and questions can arise at multiple levels, 
from local clarification requests to 
issues at discourse-level 
. In addition, discourse structure is shaped not only by 
content but also by interactional phenomena such as turn-taking, speaker intentions, and shared situational context, all of which complicate identifying a well-defined set of QUDs or PQs in spontaneous data.

To address this challenge, \citet{wu2024questions} propose an empirical approach 
where large language models (LLMs) are used to generate PQs anchored in textual context. They show that, in monologic data such as news articles and TED talks, the salience of a generated question correlates with whether it is subsequently answered in the text, suggesting that discourse is organised in anticipation of the recipient's expectations.

However, it remains unclear to what extent these findings extend to dialogue. Unlike monologic discourse, dialogue is jointly constructed 
by multiple participants, and its structure emerges from interactional negotiation rather than a single authorial plan. 
To investigate whether the QUD/PQ 
generalises to naturalistic conversation
, we apply the methodology of \citet{wu2024questions} to English dialogues exhibiting varying 
interactional organisation, and examine whether the relationship between question salience and 
later resolution persists in these 
settings. By testing this 
method on naturalistic dialogue (ND), we identify several 
challenges for a QUD-driven approach to discourse structure in conversation. We also release a 
corpus of 7,124 generated questions, each annotated for salience and answerability by two annotators\footnote{\url{https://gitlab.inria.fr/adecker/dialoguepredictability.git}}. 

The remainder of the paper is organised as follows. \Cref{sec:background} reviews the theoretical and empirical work underpinning our study, \cref{sec:corpus,sec:method} present the corpus and methodology, and \cref{sec:released_corpus} describes the released dataset. We then 
discuss the results in \cref{sec:analysis} before concluding in \cref{sec:ccl}.

\section{Background}\label{sec:background}

    Questions have long played a central role in 
    theories of discourse and dialogue structure. In particular, the QUD framework models discourse as evolving around explicit or implicit questions that guide interpretation and production. Originally developed to account for conversational coherence~\citep{van1995discourse,ginzburg1995resolving,roberts1996information}, it has 
    recently attracted interest in NLP 
    for operationalising discourse structure. 

    \subsection{Questions and Discourse structure}

        \paragraph{QUDs as predictive tools.} 

        QUD-driven approaches represent discourse as a hierarchy of questions, where each contribution either (partially) answers a question, introduces a sub-question, or shifts the focus of the interaction~\citep{van1995discourse,roberts2012information,ginzburg2012interactive}. Traditionally, QUDs were used retrospectively: analysts inferred which question must have been under discussion to make an utterance coherent. More recent work instead 
        argues that discourse participants actively 
        anticipate forthcoming QUDs and use them to guide interpretation and production. 

        A growing body of evidence supports that QUDs are a cognitively relevant construct 
        used when producing and interpreting discourse. Eye-tracking experiments by \citet{Clifton2012Discourse} show that comprehension is facilitated when readers' expectations 
        align with subsequent text. According to ERP (event-related brain potential) studies by \citet{Delogu2020Expectations},  
        discourse segments failing to address an expected QUD incur higher processing costs. Extending this work computationally, \citet{wu2024questions} leveraged LLMs to generate PQs in news 
        and TED talks, and found that 
        salient questions are more likely to be answered later in the discourse. This suggests that authors structure discourse around 
        recipients' expectations. 

        Empirical evidence from dialogue remains 
        limited. \citet{Kehler2017Evaluating} showed that grammatical context influences the QUDs evoked during dialogue interpretation, although their experiments relied on controlled completion tasks. \citet{Westera2019Asking} proposed a methodology for collecting actual and potential QUDs in ND, while \citet{mulligan2025analyzing} reported 
        moderate agreement between annotators when inferring QUDs from interviews. Together, these studies highlight both the relevance of QUDs and the large space of plausible questions associated with a 
        context. However, because existing work has focused on relatively structured interactions, it remains unclear how well the framework generalises to less organised forms of conversation. We address this question by applying the methodology of \citet{wu2024questions} to a diverse set of ND.

        \paragraph{Operationalising QUDs through question \textit{salience} and \textit{answerability}.}\label{sec:qsalience}

        \citet{wu2024questions} introduced the first computational framework designed to test whether salient questions are more likely to be answered in subsequent discourse. Working with English news articles~\citep{ko-etal-2022-discourse,huang-etal-2024-embrace} and TED talks~\citep{westera-etal-2020-ted}, they considered questions evoked at a sentence $S_k$ given the preceding context $S_1,\ldots,S_{k-1}$ and evaluated whether these questions were answered in the remaining text.

        \begin{figure*}[h!]
            \begin{tabular}{p{.95\textwidth}}
                \toprule
                $[1]$ Amid skepticism that Russia's war in Chechnya can be ended across a negotiating table, peace talks were set to resume Wednesday in neighboring Ingushetia. \textbf{$[2]$ The scheduled resumption of talks in the town of Sleptsovsk came two days after agreement on a limited cease-fire, calling for both sides to stop using heavy artillery Tuesday.} $[3]$ They also agreed in principle to work out a mechanism for exchanging prisoners of war and the dead. \\
                \midrule
                $[Q1]$ What other progress has been made towards peace recently? \\
                \bottomrule
            \end{tabular}

            \vspace{1em}

            \begin{itemize}[noitemsep,topsep=0pt]
                \item Is $[Q1]$ salient given the context of $[1]$ and the \textbf{anchor $[2]$}?
                \begin{itemize}
                    \item[] $\rightarrow$ Yes, it elaborates on what is being done to establish peace in the region.
                \end{itemize}
                \item Is it answered in the subsequent text?
                \begin{itemize}
                    \item[] $\rightarrow$ Yes, in $[3]$.
                \end{itemize}
            \end{itemize}

            \caption{Example from \citet{wu2024questions} of a generated question based on a piece of news text.}
            \label{fig:QSalience_example}
        \end{figure*}

        To operationalise this idea, they leveraged LLMs to generate questions anchored in $S_k$ and annotated each question for \textit{salience} -- its relevance at that point in the discourse -- and \textit{answerability} -- the extent to which it is addressed later in the text. \Cref{fig:QSalience_example} provides an example from their dataset. Spearman correlation between these two variables revealed strong positive relationships ($\rho=.59$ for news and $\rho=.74$ for TED talks). They also 
        introduced QSalience, a Mistral-7B-Instruct model fine-tuned to predict question salience from context.
       
        These findings 
        provide a first large-scale operationalisation of QUDs as a predictive model of discourse structure. However, because the study focuses on monologic genres, it remains unclear whether the same relationship holds in dialogue.

    \subsection{The Multifaceted Nature of Dialogue}

        Dialogue encompasses interaction types ranging from highly structured exchanges to informal conversations. Such variation 
        affects 
        relevance of PQ and the likelihood that they will be addressed. Questions 
        in 
        interviews or meetings are 
        constrained by shared goals, whereas in casual conversation they seem more opportunistic and less predictable. 
        The number of participants 
        influences conversational organisation. Duologues, \ie{} dialogues between two participants, 
        exhibit more stable interactional trajectories, whereas multilogues, \ie{} dialogues with three or more participants, introduce 
        complexity in turn management and discourse development~\citep{traum2003issues}. Larger conversations may 
        split into concurrent conversational threads, 
        known as the \textit{dinner party problem}~\citep{Dunbar1995SizeStructure,Dezecache2012Joke,Krems2016ConversationSize,KREMS2019ConversationsLimited} or \textit{schisming}~\citep{SimplestSystematics}. Analysing such interactions from transcripts 
        can 
        be challenging~\citep{hunter2015RFC}. 

        More fundamentally, dialogue is situated~\citep{Goodwin2000Action}. Understanding an interaction often requires access to shared knowledge, speaker intentions, interpersonal relations, and 
        the physical environment, information 
        rarely available in transcripts. Dialogue also requires participants to coordinate the interaction 
        through turn-taking~\citep{SimplestSystematics}, grounding~\citep{ClarkBrennan1991}, and participation management. Consequently, salient questions may concern not only information exchange but also interactional issues such as who should speak next or whether a contribution has been understood. 

        Together, these properties make dialogue a substantially more complex environment for QUD-based modelling than the 
        genres studied previously as they all increase the range of 
        relevant PQ and may reduce the predictability of subsequent turns.

\section{Source Corpus: Naturalistic dialogues}\label{sec:corpus}

    For our experiments we use the spoken component of the British National Corpus (BNC) in its XML format\footnote{The \href{http://hdl.handle.net/20.500.14106/2554}{XML version of the BNC} is freely available for research under the terms of the \href{https://www.natcorp.ox.ac.uk/docs/licence.html}{BNC User Licence}.}, a large collection of transcribed British English conversations 
    across interactional settings. To investigate the robustness of the PQ approach across different 
    forms of interaction, we selected 60 dialogues varying in both the number of participants and 
    expected degree of interactional organisation. In the end, we designed 6 categories around these two dimensions (\cref{tab:subgroups}). 

    \begin{table*}[ht!]
        \centering
        \begin{tabular}{p{.117\textwidth}p{.184\textwidth}cccC{.09\textwidth}C{.079\textwidth}C{.1\textwidth}}
            \toprule
            \multirow{2}{*}{\textbf{Group}} & \multirow{2}{*}{\textbf{Context}} & \multirow{2}{*}{\textbf{\# Ppts}} & \multirow{2}{*}{\textbf{Length}} & \multirow{2}{*}{\textbf{\# Turns}} & \textbf{\# Tokens / turn} & \textbf{\% Unk. spkr} & \textbf{\# Gen. questions} \\ \midrule
            Open duo & Spontaneous \newline conversations & $2$ & $40-70$ &47$\pm$8&8$\pm$13&7$\pm$22&66$\pm$11\\
            Open multi & Spontaneous \newline conversations & $>4$ & $40-90$ &58$\pm$7&7$\pm$7&4$\pm$9&82$\pm$11\\
            Org. duo & Interviews & $2$ & $\geq 40$ &107$\pm$3&17$\pm$20&0$\pm$0&155$\pm$5\\
            Org. multi & Committee \newline meetings & $>4$ & $\geq 40$ &97$\pm$12&18$\pm$22&25$\pm$21&141$\pm$18\\
            Public duo & `Ideas in action' \newline radio broadcasts & $2$ & $40-90$ &81$\pm$7&32$\pm$27&3$\pm$11&116$\pm$10\\
            Public multi & Radio broadcasts & $>4$ & $\geq 40$ &106$\pm$18&23$\pm$30&6$\pm$7&154$\pm$27\\
                & Television debates / discussions & $>4$ & $\geq 40$ &105$\pm$22&14$\pm$16&6$\pm$3&152$\pm$33\\
            \bottomrule
        \end{tabular}
        \caption{Description of the groups of dialogues in our corpus.}
        \label{tab:subgroups}
    \end{table*}

    The first two 
    are spontaneous conversations drawn from the demographically sampled portion of the BNC. 
    For the others, we distinguish organised discussions (interviews and committee meetings) from public discussions (radio and television programmes
    ) 
    because both rely on an 
    agenda but the latter are intended for an external audience. 
    
    For each category we randomly sampled ten dialogues, except for public multilogues, which were split equally between radio and television discussions. When 
    dialogues were too long for manual annotation, we extracted shorter excerpts while preserving the interactional characteristics. 
    Excerpts were truncated manually at the end of a topic and
    checked to ensure that the intended number of participants remained active. 
    Sampling constraints 
    are 
    in \cref{tab:subgroups-constraints}. 

    Descriptive statistics for the resulting corpus 
    show that spontaneous conversations are generally shorter and contain briefer turns than 
    more structured interactions
    , where participants 
    produce long explanations or arguments.

    The corpus also reflects some of the challenges associated with 
    working with ND transcripts. Most categories contain a small proportion of turns whose speaker is marked as unknown
    , and occasional speaker attribution errors were noted by annotators. 
    Approximately 2\% of tokens correspond to \texttt{<unclear>} markers, indicating stretches of speech that could not be reliably transcribed, 
    sometimes complicating the 
    annotations.

\section{Method: Automatic question generation and human annotation}\label{sec:method}

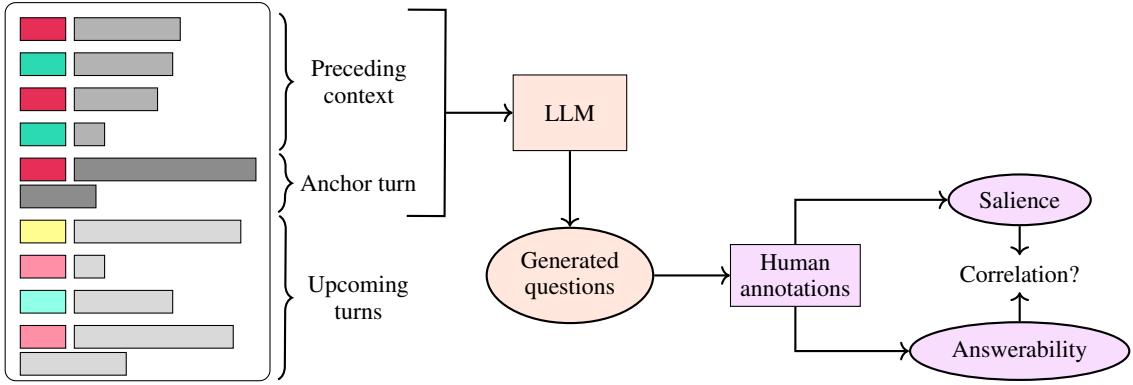
\begin{figure*}[ht!]
        \centering
        \begin{tikzpicture}[
            font=\footnotesize
        ]
            \node[rectangle, rounded corners, draw, minimum width=3.5cm, minimum height=5cm] (doc) {};
            \node[rectangle, draw, minimum width=6mm, minimum height=3mm, fill=pptred] (spk1) at ($(doc.north west) + (2mm, -2mm)$) [anchor=north west] {};
            \node[right = 1mm of spk1, anchor=west, rectangle, draw, minimum width=14mm, minimum height=3mm, fill=gray!60] (u1) {};
            \node[below = 1.5mm of spk1, anchor=north, rectangle, draw, minimum width=6mm, minimum height=3mm, fill=pptteal] (spk2) {};
            \node[right = 1mm of spk2, anchor=west, rectangle, draw, minimum width=13mm, minimum height=3mm, fill=gray!60] (u2) {};
            \node[below = 1.5mm of spk2, anchor=north, rectangle, draw, minimum width=6mm, minimum height=3mm, fill=pptred] (spk3) {};
            \node[right = 1mm of spk3, anchor=west, rectangle, draw, minimum width=11mm, minimum height=3mm, fill=gray!60] (u3) {};
            \node[below = 1.5mm of spk3, anchor=north, rectangle, draw, minimum width=6mm, minimum height=3mm, fill=pptteal] (spk4) {};
            \node[right = 1mm of spk4, anchor=west, rectangle, draw, minimum width=4mm, minimum height=3mm, fill=gray!60] (u4) {};
            \node[below = 1.5mm of spk4, anchor=north, rectangle, draw, minimum width=6mm, minimum height=3mm, fill=pptred] (spk5) {};
            \node[right = 1mm of spk5, anchor=west, rectangle, draw, minimum width=24mm, minimum height=3mm, fill=gray!90] (u5) {};
            \node[below = 2mm of spk5.south west, anchor=west, rectangle, draw, minimum width=10mm, minimum height=3mm, fill=gray!90] (u5bis) {};
            \node[below = 1.5mm of u5bis.south west, anchor=north west, rectangle, draw, minimum width=6mm, minimum height=3mm, fill=mylemon] (spk6) {};
            \node[right = 1mm of spk6, anchor=west, rectangle, draw, minimum width=22mm, minimum height=3mm, fill=gray!30] (u6) {};
            \node[below = 1.5mm of spk6, anchor=north, rectangle, draw, minimum width=6mm, minimum height=3mm, fill=mystrawberry] (spk7) {};
            \node[right = 1mm of spk7, anchor=west, rectangle, draw, minimum width=4mm, minimum height=3mm, fill=gray!30] (u7) {};
            \node[below = 1.5mm of spk7, anchor=north, rectangle, draw, minimum width=6mm, minimum height=3mm, fill=myteal] (spk8) {};
            \node[right = 1mm of spk8, anchor=west, rectangle, draw, minimum width=13mm, minimum height=3mm, fill=gray!30] (u8) {};
            \node[below = 1.5mm of spk8, anchor=north, rectangle, draw, minimum width=6mm, minimum height=3mm, fill=mystrawberry] (spk9) {};
            \node[right = 1mm of spk9, anchor=west, rectangle, draw, minimum width=21mm, minimum height=3mm, fill=gray!30] (u9) {};
            \node[below = 2mm of spk9.south west, anchor=west, rectangle, draw, minimum width=14mm, minimum height=3mm, fill=gray!30] (u9bis) {};

            \draw[decorate,decoration={brace,amplitude=5pt},thick] ($(u1.north east) + (13mm,.5mm)$) -- ($(u4.south east) + (23mm,-.5mm)$) node[midway,xshift=30pt,align=center,font=\footnotesize] (Cp) {Preceding\\context};
            \draw[decorate,decoration={brace,amplitude=5pt},thick] ($(u5.north east) + (3mm,.5mm)$) -- ($(u5bis.south east) + (24mm,-.5mm)$) node[midway,xshift=30pt,minimum width=18mm] (At) {\footnotesize Anchor turn};
            \draw[decorate,decoration={brace,amplitude=5pt},thick] ($(u6.north east) + (5mm,.5mm)$) -- ($(u9bis.south east) + (20mm,-.5mm)$) node[midway,xshift=30pt,minimum width=10mm,align = center, font = \footnotesize] (Ut) {Upcoming\\turns};

            \draw [thick] ($(u1.north east) + (30mm,1mm)$) to [square left brace] ($(u5bis.south east) + (41mm,-1mm)$);
            \node[rectangle, draw, fill=myorange!30, minimum width=15mm, minimum height=10mm, align=center, font=\footnotesize] (llm) at ($(u1.north east)!.5!(u5bis.south east) + (5.7cm,0)$) {LLM};
            \node[below = of llm, thick, ellipse, align = center, draw, fill = myorange!30] (qgen) {Generated\\questions};
            \node[right = of qgen, rectangle, draw, fill=myvioline!30, align=center] (annotations) {Human\\annotations};
            \node[thick, ellipse, align = center, draw, fill=myvioline!30, anchor=center] (sal) at ($(annotations.east) + (21mm, 10mm)$) {Salience};
            \node[thick, ellipse, align = center, draw, fill=myvioline!30, anchor=center] (answ) at ($(annotations.east) + (21mm,-10mm)$) {Answerability};
            \node[] at ($(sal.south)!.5!(answ.north)$) (correlation) {Correlation?};

            \draw[->, thick] ($(llm) + (-16.5mm,0)$) -- (llm);
            \draw[->, thick] (llm) -- (qgen);
            \draw[->, thick] (qgen) -- (annotations);
            \draw[->, thick] (annotations) |- (sal);
            \draw[->, thick] (annotations) |- (answ);
            \draw[->, thick] (sal) -- (correlation);
            \draw[->, thick] (answ) -- (correlation);
        \end{tikzpicture}
        \caption{Framework for correlating question salience and answerability, adapted from \citet{wu2024questions}.}
        \label{fig:qsalience-framework}
    \end{figure*}

    We adapt the framework of \citet{wu2024questions} to dialogue. Given a dialogue composed of utterances $U_1,\ldots,U_n$, we consider an anchor utterance $U_k$ and its preceding context $U_1,\ldots,U_{k-1}$. We generate PQs evoked at that point in the interaction and examine whether their salience is correlated with the extent to which they are answered in the subsequent dialogue $U_{k+1},\ldots,U_n$ (\cref{fig:qsalience-framework}). 

    The methodology consists of three steps: (i) generating questions from dialogue context using an LLM, (ii) manually annotating the resulting questions for salience and answerability, and (iii) measuring the correlation between these two variables.

    \subsection{Automatic question generation}\label{sec:method_questions_generation}
        
        Following \citet{wu2024questions}, we generate three questions per anchor utterance, selecting every second utterance from the fourth onward and pairing it with its preceding context.
        We tested three open-source LLMs for question generation: Gemma-3-4B~\citep{gemmateam2024gemmaopenmodelsbased}, Llama3-ChatQA-8B~\citep{liu2024chatqasurpassinggpt4conversational}, and Mistral-7B~\citep{jiang2023mistral7b}. Using a dialogue-adapted 
        prompt 
        from \citet{wu2024questions} (\cref{figapp:prompt}), we conducted a pilot evaluation of the generated questions and selected Mistral for all subsequent experiments. In particular, we dismissed Llama because of the blatantly sexist questions generated by it as shown in \cref{fig:ex_sexism}. LLMs are known to be biased~\citep{gehman-etal-2020-realtoxicityprompts}, and this example illustrates how these biases can manifest in the generated questions. 
        After a first small batch of annotations (about 150 questions) carried out by one person, we selected Mistral rather than Gemma because the generated questions were more salient in average and more often valid. We acknowledge that this sample is too small to draw any strong conclusions but it seemed to be a reasonable choice given the time constraints of the project. We chose to dedicate more time to the final stage of the annotation process in order to compare different types of dialogues on a large enough sample to draw meaningful conclusions. Investigating the impact of the choice of LLM on the quality of the generated questions is an interesting avenue for future work. 

        \begin{figure*}
            \begin{itemize}
            \setlength\itemsep{-.2em}
            \item[] {\color{pptpurple}PS0BA$_1$}: Have you been?
            \item[] {\color{pptorange}PS0BG$_2$}: I got slightly lost.
            \item[] {\color{pptorange}PS0BG$_3$}: Well I sort of went back down the  <unclear> Road cos there was a traffic jam  <pause> and er I went down  <pause> oh sorry mate  <pause> went down a road like which I thought was a good short cut, and they've got a lot of these roads and they've put like pavements across the end of them with bollards, and I went back and then there was a traffic jam and I got stuck because they were unloading a lorry  <pause> so I've been er basically pissed about.<laugh><pause>
            \item[] {\color{pptpurple}PS0BA$_4$}: \textbf{Yeah she a bit <unclear> now ain't she?}
            \begin{itemize}
                \item[] $\rightarrow$ \textbf{Generated question:} Why is there uncertainty whether the female in question is pregnant?
                \item[] $\rightarrow$ \textbf{Generated question:} How does the speaker know that the female in question isn't pregnant?
            \end{itemize}
            \end{itemize}
            \caption{Examples of sexist questions generated by Llama3-ChatQA-8B. \textit{Beginning of the example KC6 -- 055905.}}\label{fig:ex_sexism}
        \end{figure*}

    \subsection{Guidelines adaptations for dialogue}\label{sec:method_guidelines}
    
        We adapted the annotation scheme of \citet{wu2024questions} to account for dialogue-specific phenomena. 

        \paragraph{Salience.} It measures the extent to which answering a question would contribute to understanding the dialogue at a given point. We retain the original scale while introducing three dialogue-specific categories (italicised below), leading to scores:

        \begin{itemize}
            \setlength\itemsep{0em}
            \item[0:] Invalid question, \ie{} it contains a grammatical error, contradicts the preceding context or is unrelated to the anchor turn;
            \item[1:] Question unrelated to the preceding context;
            \item[2:] \textit{Question related to issues in the transcription of the anchor turn};
            \item[3:] Question related to the preceding context but answering it is not useful, it may be fully answered already;
            \item[4:] \textit{Relevant question but the participants clearly already know the answer based on their common ground or shared context};
            \item[5:] Relevant but not essential question;
            \item[6:] Relevant and somewhat useful question;
            \item[7:] \textit{The question is a paraphrase of the anchor 
            };
            \item[8:] Relevant question, 
            must be answered.
        \end{itemize}
        
        The final scheme emerged through several pilot annotation rounds in which alternative dialogue-specific categories were tested and refined. 

        \paragraph{Answerability.} 
        It measures the extent to which a question is addressed after the anchor. 
        Unlike \citet{wu2024questions}, we do not distinguish questions already answered by the preceding context, 
        since information may 
        recur 
        later in dialogue. We therefore use a three-point scale with the following scores:
        
        \begin{itemize}
            \setlength\itemsep{0em}
            \item[0:] Question not answered after the anchor turn;
            \item[1:] Question partially answered or the participants explicitly state that they do not know the answer after the anchor turn;
            \item[2:] Question fully answered after the anchor turn.
        \end{itemize}

    \subsection{Annotations}\label{sec:method_annotations}
        
        Each question was independently annotated for salience and answerability by two annotators. Annotator 1 corresponds to a single researcher with training in linguistics and dialogue analysis, while annotator 2 aggregates annotations produced by several researchers with varying levels of expertise. Despite the aggregation for Annotator 2, one person carried out the annotation of most of the files -- 77\% for salience, and 95\% for answerability -- and 7 other annotators covered the rest of the files, 6 for salience and 1 for answerability.
        All annotators were high-proficiency L2 speakers of English. 

        Annotations were collected through a custom web-based interface\footnote{\href{https://dialoguequds-f1c8a6.gitlabpages.inria.fr}{Salience} and \href{https://dialoguequds-f1c8a6.gitlabpages.inria.fr/answerability}{Answerability} interfaces.} that highlighted the anchor utterance.
        For salience annotation, only the dialogue preceding the anchor 
        was displayed; 
        for answerability
        , annotators had access to the full dialogue. 

        \paragraph{Agreement.} Given the inherently subjective nature of both salience and answerability annotation, we evaluate inter-annotator agreement using multiple metrics. We report Krippendorff's Alpha~\citep{krippendorff2011computing}\footnote{Computed with \citeposs{castro-2017-fast-krippendorff} Python implementation.} for comparability with \citet{wu2024questions}, as well as Cohen's Kappa and both linear and quadratic weighted Kappa~\citep{cohen1968weighted} to account for the ordinal structure in the annotation scales. Indeed, given the gradation of the scores, we consider that a disagreement between two adjacent scores is less severe than a disagreement between two scores far apart. Weighted $\kappa$ allows us to take this into account in our agreement analysis, where instead of penalising all disagreements with a weight of 1 as in the case of Cohen's Kappa, we assign a weight to each disagreement based on the distance between the two scores. Formally, the weight when comparing two scores $i$ and $j$ is defined as $w_{i,j} = |j-i|$ for linear weighted Kappa and $w_{i,j} = (j-i)^2$ for quadratic weighted Kappa. We discuss the agreement values in \cref{sec:salience,sec:answerability}. 

        \paragraph{Aggregation.} Given the observed variability in salience annotations, we adopt a \textit{perspectivist} strategy in our analysis~\citep{frenda2025perspectivist}. Rather than collapsing annotations into a single gold label via averaging, we retain both annotators' perspectives. 
        We also construct a restricted `gold' subset 
        of questions for which both annotators assigned identical salience scores. This allows us to study the general patterns 
        and investigate the nature of the disagreements which we consider to be a natural reflection of the complexity of the task and of the nature of dialogue itself, 
        perceived differently by its different participants. 
        Regarding answerability, we 
        compute the mean of the two annotators' scores to obtain a single continuous estimate. 
        The resulting aggregation effectively yields a finer-grained scale while preserving ordinal consistency.

    \begin{figure*}[ht]
        \centering

        \begin{itemize}
            \setlength\itemsep{-.2em}
            \item[] {\color{pptred}\textsc{PS0EH\_0}}: I was going to tell you about Margaret
            \item[] {\color{pptteal}\textsc{PS0ED\_1}}: Oh yeah, Margaret and Snowy, yeah
            \item[] {\color{pptred}\textsc{PS0EH\_2}}: they went to this Howstead Abbey
            \item[] {\color{pptteal}\textsc{PS0ED\_3}}: Yeah
            \item[] {\color{pptred}\textsc{PS0EH\_4}}: \textbf{because they've got a season ticket and they got to feed the ducks and}
        \end{itemize}
        
        \vspace{1em}

        \begin{tabularx}{\textwidth}{ccccccX}
            \toprule
            ID$_{sheet}$ & ID$_{BNC}$ & Group & ID$_Q$ & Cutoff & ID$_{ann2}$ & Question \\ \midrule
            4 & KCE -- 029701 & Open multi & 0 & 4 & E & \textbf{1. Do Margaret and Snowy visit Howstead Abbey every season?} \\
            \bottomrule
        \end{tabularx}

        \begin{tabularx}{\textwidth}{ccccXccccc}
            \toprule
            S$_{1}$ & S$_{2}$ & S$_{G}$ & Comm$_{S1}$ & Comm$_{S2}$ & A$_{1}$ & A$_{2}$ & A$_{G}$ & Comm$_{A1}$ & Comm$_{A2}$ \\ \midrule
            5 & 5 & 5 & / & \textit{relevant but not essential to the conversation} & 0 & 0 & 0 & / & / \\
            \bottomrule
        \end{tabularx}
        \caption{Example of a row in our corpus together with an extract of the dialogue it is based on, with the anchor turn and the associated question in bold. S$_i$ and A$_i$ indicate the salience and answerability scores for annotator $i$ with $i=G$ for the gold scores. \textit{Comm} stand for the associated comments. \textit{Beginning of the dialogue KCE -- 029701.}} 
        \label{fig:example}
    \end{figure*}
        
\section{Released dataset: questions and annotations}\label{sec:released_corpus}

    We release the 7,124 questions generated by 
    Mistral-7B, together with 
    salience and answerability annotations, as well as some natural language explanations for some of the questions. Due to the licence of the BNC, we cannot include the dialogues 
    but we provide the metadata for each question, 
    allowing reconstruction of the dialogues from the BNC. We 
    include 
    code to do so on a GitLab repository.
    We release our corpus as a single file where each row corresponds to a question and includes the following information: 
    its dialogue, the anchor turn, the question, the two salience scores, the gold score when the two annotators agreed, comments from the annotators for the salience, the two answerability scores, the gold score, comments from the annotators for the answerability. \Cref{fig:example} provides an example of a row in our corpus together with an extract of the dialogue it is based on. 

\section{Results and Analysis: Dialogue is less predictable than static text}\label{sec:analysis}

    We begin by analysing salience and answerability separately, focusing on annotation patterns and differences across dialogue types, before examining the relationship between the two measures through correlation analyses. 

    \begin{figure*}
        \begin{subfigure}{0.95\columnwidth}
            \centering
            \includegraphics[width=\textwidth]{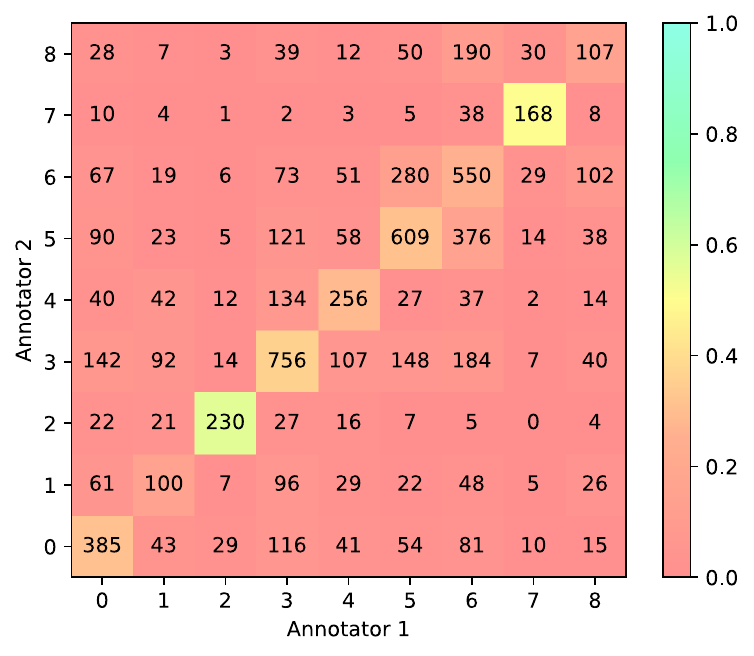}
            \caption{Salience}
            \label{fig:comparison_salience_all}
        \end{subfigure}
        \hfill
        \begin{subfigure}{0.95\columnwidth}
            \centering
            \includegraphics[width=\textwidth]{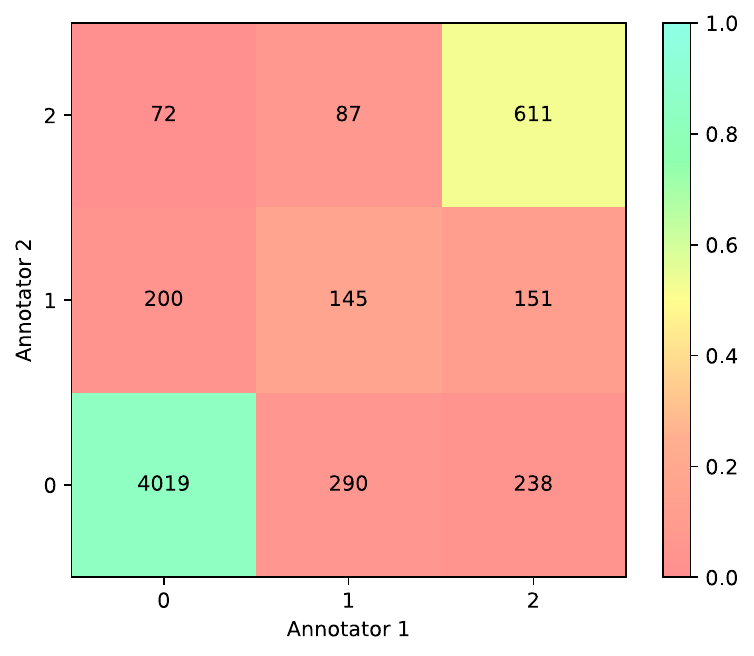}
            \caption{Answerability}
            \label{fig:comparison_answerability_all}
        \end{subfigure}
        \caption{Comparison of the annotations 
        for salience and answerability. The colour of the cells 
        corresponds to the portion of questions that received the corresponding scores from the two annotators, while the numbers indicate the actual amount of annotations for each score.}
    \end{figure*}

    \begin{table}
        \centering
        \begin{tabular}{lcccccc}
            \toprule
            & $\alpha_{QSalience}$ & $\alpha$ & \% & $\kappa$ & w$\kappa$ & w$\kappa^2$ \\
            \midrule
            \textbf{S} & .63 - .75 & .55 & .51 & .36 & .47 & .53 \\
            \textbf{A} & .80 & .65 & .80 & .53 & .62 & .68 \\
            \bottomrule
        \end{tabular}
        \caption{Inter-annotator agreement for salience and answerability annotations. $\alpha_{QSalience}$ refers to the Krippendorff's alpha reported in \citet{wu2024questions} for comparison, the other scores were computed on our dataset, with $\%$ referring to raw percentage agreement, $\kappa$ to Cohen's Kappa (CK), w$\kappa$ to linear weighted CK 
        and w$\kappa^2$ to quadratic weighted CK. 
        }
        \label{tab:agreement}
    \end{table}

    \subsection{Salience: Which questions should be answered?}\label{sec:salience}

    The salience annotation relied on a large score range. As shown in \cref{tab:agreement}, agreement is fair to moderate depending on the metric, and lower than that reported by \citet{wu2024questions}, likely due to dialogue complexity and the more fine-grained annotation scheme. Nevertheless, we considered this agreement level reasonable given the complexity and subjectivity of the task and proceeded with the analysis. Disagreements are partly attributable to transcription artefacts and the absence of non-linguistic contextual information, both of which are particularly relevant in spontaneous dialogue. 

\begin{figure}
        \centering
        \includegraphics[width=\columnwidth]{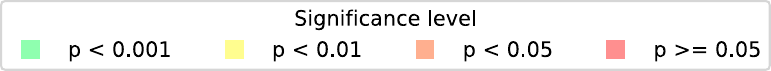}
        
        \includegraphics[width=\columnwidth]{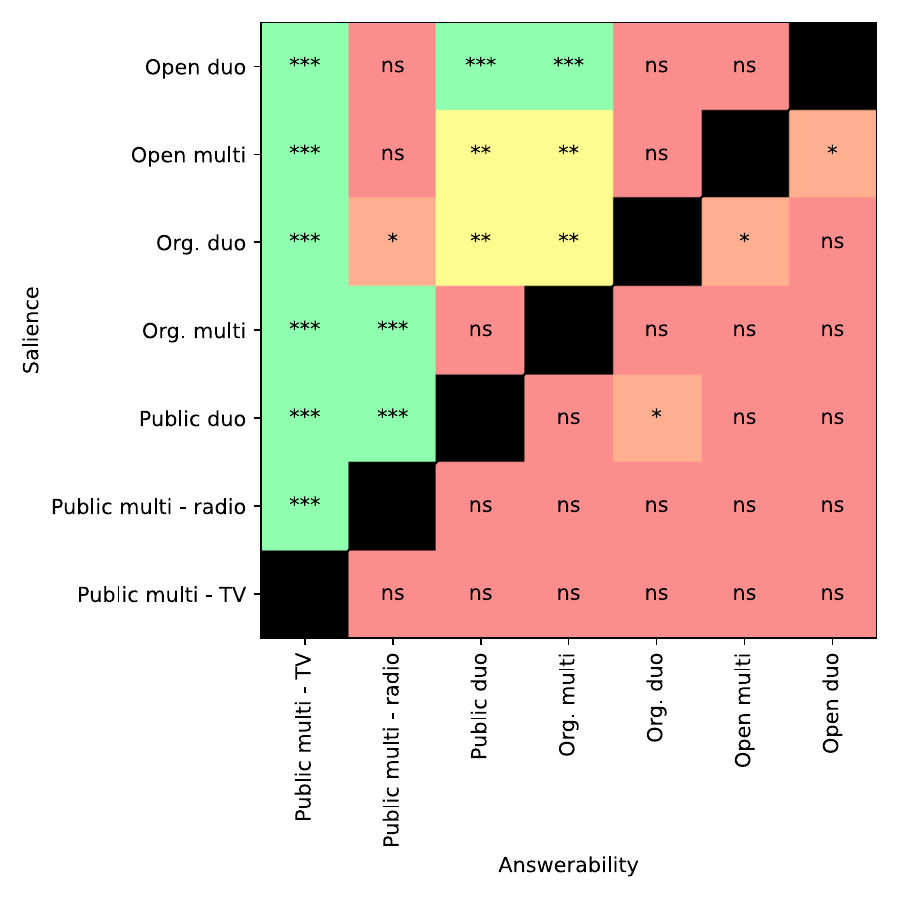}
        \caption{Representation of the significance of the difference between the agreement for each pair of dialogue groups for Salience (◤) and Answerability (◢). Colours indicate the strength of the p-value. }
        \label{fig:agreements_groups_pvalues}
    \end{figure}

    We further compute agreement separately across dialogue categories (\cref{tab:IAA_groups}) in order to assess whether the conversational structure influences the consistency of the annotations. \Cref{fig:agreements_groups_pvalues} shows that most pairs exhibit significantly different levels of inter-annotator agreement across groups, assessed using paired Wald’s z-tests comparing the difference in unweighted Cohen’s kappa, normalised by standard errors based on Fleiss’s method~\citep{Fleiss2003Statistical}. The lowest level of agreement was observed in the public television category, which lies close to the lower limit of what is considered fair agreement. This may be explained by the television context, which is intended for an external audience with access to the visual modality, whereas this was not the case for the annotators. Conversely, the highest agreement was found in radio conversations, which are intended for an audience with access only to the audio modality. 

    \begin{table}
        \centering
        \begin{tabular}{lcc}
            \toprule
            Group & S & A \\
            \midrule
            Open duo & .40 & .56 \\
            Open multi & .38 & .47 \\
            Org. duo & .38 & .56 \\
            Org. multi & .33 & .54 \\
            Public duo & .32 & .50 \\
            Public multi - radio & .42 & .50 \\
            Public multi - TV & .25 & .53 \\
            \midrule
            All & .36 & .53 \\
            \bottomrule
        \end{tabular}
        \caption{Inter-annotator (Cohen's Kappa) agreement for each dialogue group.}
        \label{tab:IAA_groups}
    \end{table}

    \Cref{fig:comparison_salience_all} shows the cross-annotation matrix for salience. Agreement is highest for Scores 2 and 7 that correspond respectively to questions targeting transcription issues, and cases where the generated question closely paraphrases the anchor utterance. Both categories are relatively constrained and therefore more consistently identifiable across annotators. In particular, Score 2 can often be detected using simple heuristics (\eg{} the presence of \texttt{<unclear>} markers in the question), which likely contributes to higher consistency. Score 7 is slightly less consensual which can be explained by the fact that the boundary between two sentences being paraphrases of each other, and being very related, is hard to define~\citep{polguere-1990}. 

    In contrast, lower agreement is observed for Scores 1 and 8. Score 1 involves judgements about whether a question is unrelated to the preceding context, a notion 
    inherently sensitive to assumptions of discourse coherence. Conversely, Score 8 requires assessing whether a question `must be answered', which depends on pragmatics 
    and access to shared context and 
    intentions, both of which are 
    unobservable in transcript-based analysis. 

    The distribution shows that annotators frequently agree that questions are at least loosely connected to the preceding context, as indicated by the faint orange square spanning Scores 3 to 8 with the exception of 7
    . This supports the idea that in conversation we tend to assume coherence and relatedness by default, which is a reasonable strategy for joint cooperation~\citep[p.45–47]{LogicandConversation}. However, the extent to which a question is deemed essential 
    to the dialogue is more 
    variable across annotators. In particular it can require a more fine-grained understanding of the dialogue, which can be difficult to achieve without access to the non-linguistic context and to the knowledge of the participants. 
    
    \begin{table}
        \centering
        \begin{tabular}{lcccc}
            \toprule
            Group & S$_1$ & S$_2$ & S$_G$ & Supp.\\ 
            \midrule
            Open duo & 3.3 & 3.3 & 3.0 & 321 \\ 
            Open multi & 3.3 & 3.2 & 3.1 & 388 \\ 
            Org. duo & 4.0 & 4.1 & 4.1 & 742 \\ 
            Org. multi & 3.8 & 3.8 & 3.6 & 598 \\ 
            Public duo & 4.8 & 4.7 & 4.8 & 535 \\ 
            Public multi - radio & 4.2 & 3.9 & 4.0 & 402 \\ 
            Public multi - TV & 4.5 & 4.7 & 4.8 & 287 \\ 
            \midrule
            All & 3.9 & 3.9 & 3.8 & 3273 \\ 
            \bottomrule
        \end{tabular}
        \caption{Average salience score per dialogue group, with Supp. indicating the number of questions in each category.}
        \label{tab:avg_salience}
    \end{table}

    When comparing salience distributions across dialogue categories, we observe systematic variation between groups. Pairwise comparisons (\cref{figapp:salience_distribution}) indicate that only a few pairs of groups display non-significant differences. Open-domain duologues and multilogues are similar together while being different from all the other groups in both annotators' distributions. Public radio multilogues 
    are similar to the organised duologues, while public television multilogues 
    are similar to the public duologues, although the statistical strength of these differences varies by annotator. When looking into the differences for the gold annotations, we observe no significant differences between the groups, 
    suggesting that disagreements between the annotators highlight more complex interactions, characteristic of the different dialogue types. This last result should be interpreted with caution, as the gold annotations represent a restricted subset of the data which limits the statistical power of the tests. 
    These results suggest that salience is sensitive to dialogue type and in particular to the number of participants, 
    duologues displaying more salient questions than multilogues (\cref{tab:avg_salience}). 
 
    \subsection{Answerability: Which questions are answered?}\label{sec:answerability}

    Answerability was graded on a small score range. As shown in \cref{tab:agreement}, agreement is higher for answerability than for salience, reflecting the lower interpretative flexibility of answerability judgements. This is confirmed by the few significant differences in terms of agreement across dialogue types as shown in \cref{fig:agreements_groups_pvalues}. 
    \Cref{fig:comparison_answerability_all} shows the corresponding annotation matrix. Agreement is highest for Score 0, which corresponds to cases where the question is not answered in the subsequent dialogue. This suggests that identifying absence of resolution is more robust than distinguishing degrees of partial resolution. Disagreements between Scores 1 and 2 can be explained by the fact that the boundary between partial and full resolution is ambiguous, particularly for complex or implicit questions. 
    Requesting a confidence score for the answerability annotation could have been useful to capture such cases, but it would also make the annotation process more complex and time-consuming. 

    \begin{table}
        \centering
        \begin{tabular}{lccc}
            \toprule
            Group & A$_1$ & A$_2$ & A$_G$ \\ 
            \midrule
            Open duo & .36 & .34 & .35 \\ 
            Open multi & .20 & .19 & .20 \\ 
            Org. duo & .47 & .40 & .44 \\ 
            Org. multi & .41 & .30 & .35 \\ 
            Public duo & .55 & .48 & .52 \\ 
            Public multi - radio & .56 & .44 & .50 \\ 
            Public multi - TV & .44 & .36 & .40 \\ 
            \midrule
            All & .42 & .35 & .38 \\ 
            \bottomrule
        \end{tabular}
        \caption{Average answerability score per 
        group.}
        \label{tab:avg_answerability}
    \end{table}


    Pairwise statistical tests (\cref{figapp:answerability_distribution}) indicate fewer significant differences between groups, although 
    variation remains. 
    Particularly, open-domain multilogues are significantly different from all other 
    types, and organised multilogues are different from all 
    types except open-domain duologues and public television multilogues. 
    This suggests that answerability is less sensitive to dialogue type than salience. However, the low average answerability scores across \textit{all} groups indicates that most questions are not 
    answered in the subsequent dialogue (\cref{tab:avg_answerability}).

    \subsection{Dialogue predictability}\label{sec:correlation}

    \begin{figure*}
        \centering
        \includegraphics[width=.8\textwidth]{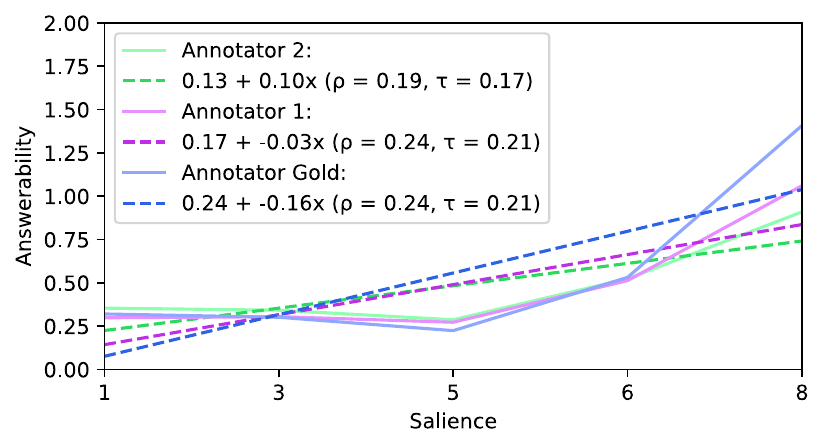}
        \caption{Correlation between salience and answerability scores ($\text{all p-values} \ll .001$). The Salience scores were mapped to a 1-5 scale for the computation; the original labels are displayed here for readability purposes. \textit{Legend in the same order as the dotted plots at 0.}}
        \label{fig:salience_answerability_correlation}
    \end{figure*}

    We compute the correlation between salience and answerability to evaluate whether more salient questions are more likely to be answered in the subsequent dialogue turns. Following \citet{wu2024questions}, we use Spearman's rank correlation coefficient to account for the ordinal nature of the annotation scales. We additionally report Kendall's $\tau$, which is less sensitive to ties and therefore suitable given the discrete label space of answerability. To retain an ordinal, monotone ordering, we compute the correlation only on questions whose salience was 1, 3, 4, 6, or 8. 
    Correlations are computed separately for each annotator as well as for aggregated (`gold') annotations. Results are shown in \cref{fig:salience_answerability_correlation}, which reports mean answerability per salience level along with regression lines, and the correlation coefficients in the legend. The results with the full scale are available in \cref{figapp:correlation}. 

    Across all settings, we observe a positive correlation between salience and answerability. This finding replicates the general trend reported by \citet{wu2024questions} in monologic data. However, the magnitude of the correlation is two to three times lower than that reported for news and TED talks ($\rho = 0.59$--$0.74$). While the model we used to generate the questions is different from \citeposs{wu2024questions} one, correlation reflects the relationship between salience and answerability. Thus, the technical ability of the model to produce relevant questions is reflected by the salience distribution rather than the salience-answerability correlation. The strength of the p-value in our case indicates that the amount of questions in each category is sufficient, indeed suggesting weaker alignment between salience and downstream resolution in conversational settings. Nevertheless, expanding this work with different models would validate this hypothesis. This reduction in correlation is consistent with the increased variability of dialogue structure, where multiple interacting factors -- including turn-taking dynamics, speaker intentions, and shared situational context -- may affect whether salient questions are explicitly addressed. 

    When examining answerability conditioned on salience levels (\cref{figapp:avg_answerability_per_salience}), we observe relatively homogeneous answerability for salience scores below 6, with the exception of Score 2. Higher salience levels (6--8) are associated with significantly higher answerability. Score 2 consistently shows lower answerability across conditions, which is expected given that most of these questions ask for clarifications about the transcriptions and are 
    unlikely to be answered in the dialogue. 

    Finally, we examine whether dialogue type affects the salience--answerability relationship (\cref{figapp:correlation_diff_groups}). Most differences in correlation are not statistically significant with the exception of open-domain multilogues, which exhibit the lowest $\rho$ coefficient. Public multilogues from television, which have intermediate $\rho$ values for both annotators, also differ significantly from most other groups; only open-domain duologues and organised multilogues show similar correlations. When ranking groups by their $\rho$ coefficients, the two annotators produce nearly identical orderings, with only two groups swapping positions, suggesting that group differences are relatively consistent across annotators. In contrast, the ranking obtained from the gold annotations differs substantially, with only the two highest-ranked groups retaining their positions. This again suggests that annotator disagreements capture meaningful differences in the dialogues. 
    



    % ...


\section{Conclusion}\label{sec:ccl}

We investigated whether the relationship between question salience and answerability observed in monologic discourse also holds in dialogue. Adapting the framework of \citet{wu2024questions}, we applied it to 60 dialogues from the BNC representing a range of interactional settings. 

Our results show a positive correlation between salience and answerability: questions judged to be more salient are more likely to be addressed later in the conversation. However, this relationship is substantially weaker than that reported for news articles and TED talks, suggesting that the structure of dialogue is less predictable than that of monologic discourse. We also show that this relationship is much weaker in open-domain multilogues. Furthermore, the reduced variability and cross-type consistency in the gold annotations suggest that annotator disagreements may reflect challenges that are specific to certain dialogue types. 

This study offers one of the first large-scale operationalisations of QUD-driven approaches for modelling ND. The moderate agreement obtained for salience annotation underscores the complexity of the task and the need to account for annotator perspective when studying dialogue structure. 

Future work should investigate the temporal dynamics of question resolution, including when questions are answered and by whom. A promising direction is developing automatic dialogue salience prediction models, enabling large-scale analysis of interactional structure. 

\clearpage
\section*{Limitations}\label{sec:limitations}

Several limitations should be considered when interpreting our results. First, the study is restricted to English dialogues from the spoken BNC, collected in the early 1990s. Although it remains a rich source of naturalistic conversation, the data reflects a specific linguistic, cultural, and technological context. In addition, some transcripts contain missing, unclear, or misattributed speech, which occasionally complicated question generation and annotation. However, such imperfections are typical of naturalistic conversational data and need not be treated purely as a corpus artefact. 

Second, our operationalisation of PQs relies on questions generated by a single LLM. Different models may produce different questions from the same dialogue context. More broadly, prior work indicates that human agreement on generating or inferring QUDs is only moderate. Accordingly, our findings should be viewed as evidence for one plausible operationalisation of discourse structure rather than an exhaustive account of all possible QUDs evoked by a dialogue. 


Finally, the study raises a methodological challenge for dialogue research: annotators were external observers with access only to transcripts and limited contextual information. Estimating salience often required reconstructing participants’ assumptions, intentions, and shared knowledge. While this is common in dialogue analysis, our results suggest that such reconstructions can vary substantially across observers, even when annotators share similar cultural and linguistic backgrounds. Expanding the diversity of annotators may reveal even greater variation, reinforcing the need to consider perspectivist approaches to dialogue annotation. 

\section*{Ethical Considerations}\label{sec:ethics}

First, generative AI tools were used only to improve the linguistic quality of this work. They were not used to generate scientific content, formulate hypotheses, design experiments, or interpret results. References were selected and added manually by the authors. Where possible, we provide direct links to the cited papers or publisher webpages. References labelled as ArXiv preprints correspond, to the best of our knowledge, to work that had not yet been published in a peer-reviewed venue at the time of writing. 

Second, the study uses data from the spoken British National Corpus, whose collection procedures included participant consent and pseudonymisation of the released transcripts. However, the recordings were made several decades ago, and it is unlikely participants anticipated all contemporary uses of conversational corpora, including large-scale computational analysis at the conversation scale. Although identifying information was removed, pseudonymisation does not guarantee anonymity; prior work suggests individuals may remain identifiable from contextual information alone~\citep{amblard2014impossibilite}. In addition, some full names can still appear in the corpus. For this reason, the dialogues should be treated as situated historical interactions rather than timeless records of participants' beliefs. The views expressed reflect specific moments, contexts, and social circumstances, and may not align with participants' current opinions. While we encourage reuse of existing corpora~--~both to reduce the cost of data collection and because it places demands on participants~--~analyses and reuse should keep the historical and social context of the data as a central consideration.

Finally, although LLMs were used to generate potential questions, all analyses reported here are based on human annotation. Future work may explore automatic salience prediction, but we view extensive manual annotation as essential for understanding the task, identifying its challenges, and providing a reliable foundation for developing and evaluating automatic methods.

\section*{Acknowledgments}
We deeply thank Clémentine Bleuze, Marie Cousin, Rémi de Vergnette, Valentin D. Richard, Vincent Tourneur, and Yoann Coudert-Osmont who contributed to the annotations of our dataset. Special thanks to Junyi Jessy Li for her presentation at CLASP (Gothenburg, Sweden) in June 2025 and the enriching subsequent discussion that led to the idea for this paper.

\bibliography{custom}


\appendix

\setcounter{table}{0}
\renewcommand{\thetable}{A\arabic{table}}
\renewcommand{\thefigure}{A\arabic{figure}}

\section{Corpus selection}\label{secapp:corpus_selection}

\Cref{tab:subgroups-constraints} describes the explicit constraints we applied to the BNC metadata when selecting the dialogues for our corpus. The open-domain conversations were drawn from the `spontaneous conversation' category, which includes daily life conversations that people were asked to record during a week. The rest of the dialogues come from a part of the corpus that put together specific types of conversations. The organised conversations include interviews done in the context of oral history projects for the duologues, and meetings that were not classified as \textit{leisure} for the multilogues. Finally for the public conversations we selected broadcast discussions. We also specified the number of participants, exactly 2 for the duologues and 5 or more for the multilogues to ensure a clear distinction between the two types of conversations. Finally, we specified the length of the dialogues to ensure that they were long enough to contain a sufficient number of questions and answers, but not too long to be too difficult to annotate. Some of the categories did not contain dialogues that were short enough so we looked for longer ones that we manually cut at the end of a topic. For the multilogues we also confirmed that the number of participants was already at least 5 in the part of the dialogue we kept. 

\begin{table*}[b!]
    \centering
    \begin{tabular}{lp{.55\textwidth}ccc}
        \toprule
        \textbf{Group} & \textbf{Metadata constraints} & \textbf{\# Ppts} & \textbf{Length} & \textbf{\#} \\ \midrule
        Open duo & classification:~spontaneous conversation & $2$ & $40-70$ & 10 \\
        Open multi & classification:~spontaneous conversation & $>4$ & $40-90$ & 10 \\
        \midrule
        Org. duo & classification:~interview oral history & $2$ & $\geq 40$ & 10 \\
        Org. multi & classification:~meeting, subcategory~$\neq$~Leisure, \newline activity:~does not contain `informal' & \multirow{2}{*}{$>4$} & \multirow{2}{*}{$\geq 40$} & \multirow{2}{*}{10} \\
        \midrule
        Public duo & classification:~broadcast discussion & $2$ & $40-90$ & 10 \\
        Public multi & classification:~broadcast discussion, title:~contains `radio' & $>4$ & $\geq 40$ & 5 \\
            & classification:~broadcast discussion, title:~contains \newline `television' & \multirow{2}{*}{$>4$} & \multirow{2}{*}{$\geq 40$} & \multirow{2}{*}{5} \\
        \bottomrule
    \end{tabular}
    \caption{Constraints for the random sampling of our corpus.}
    \label{tab:subgroups-constraints}
\end{table*}


\section{Prompt for question generation}

    We adapted the prompt used by \citet{wu2024questions} to generate questions in a dialogue setting. As shown in \cref{figapp:prompt}, we explained the format of the utterances in the prompt, and indicated that special tokens should be interpreted as extra information that can help understand the dialogue better, but are not part of the utterance itself. We also instructed the model to generate questions as if it `were a participant of the dialogue'. We made this choice to reduce the number of questions the participants clearly already know the answer to. We finished the prompt by specifying the format of the output. 

    \begin{figure}[h!]
        \texttt{The following is a dialogue where each line is in the format <speaker>: <utterance>. All the <special tokens> indicate extra linguistic information, such as pauses, vocalizations, etc. You should interpret them as extra information that can help you understand the dialogue better, but they are not part of the utterance itself.}

        \vspace{1em}

        \texttt{<conversation up to the anchor turn>}

        \vspace{1em}

        \texttt{After reading the previous dialogue, ask 3 questions about a part of the last utterance "<anchor utterance>" that you would be curious about if you were a participant of the dialogue and which you don’t have an answer for.}

        \texttt{Generate only the questions, one per line, without any additional text or explanation.}

        \caption{Prompt used for question generation.}
        \label{figapp:prompt}
    \end{figure}


\section{Annotations}

    This section provides additional details on the annotation score distributions across annotators and dialogue groups. In particular we report the average scores for each annotator and the gold scores, as well as results of significance tests for pairwise comparisons between dialogue groups. \Cref{secapp:grades} reports the average salience and answerability scores, while \cref{secapp:predictability} discusses the relationship between salience and answerability.

    \subsection{Grades}\label{secapp:grades}

\begin{figure*}[h]
        \centering
        \begin{subfigure}{.85\columnwidth}
            \centering
            \includegraphics[width=\textwidth]{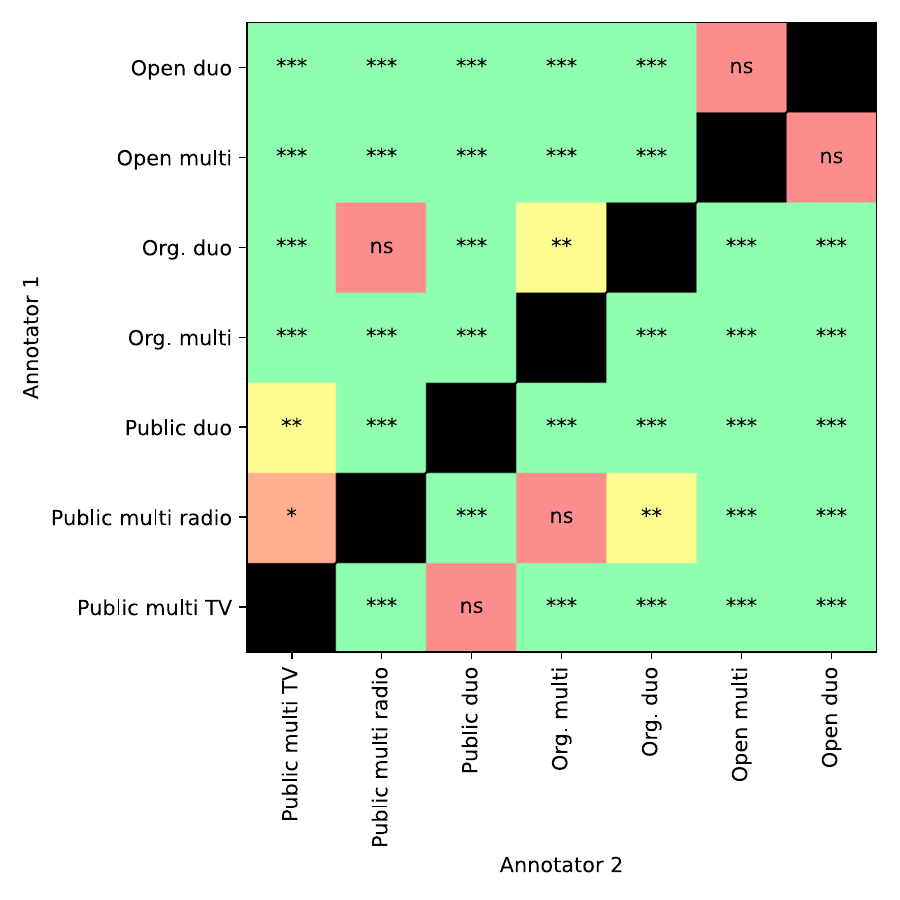}
            \caption{Annotators 1 (◤) and 2  (◢).}
            \label{figapp:salience_distribution_annotators}
        \end{subfigure}
        \hfill
        \begin{subfigure}[b]{.3\columnwidth}
            \centering
            \includegraphics[width=\textwidth]{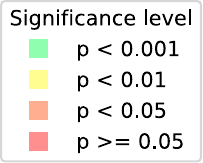}
            \vspace{8em}
        \end{subfigure}
        \hfill
        \begin{subfigure}{.85\columnwidth}
            \centering
            \includegraphics[width=\textwidth]{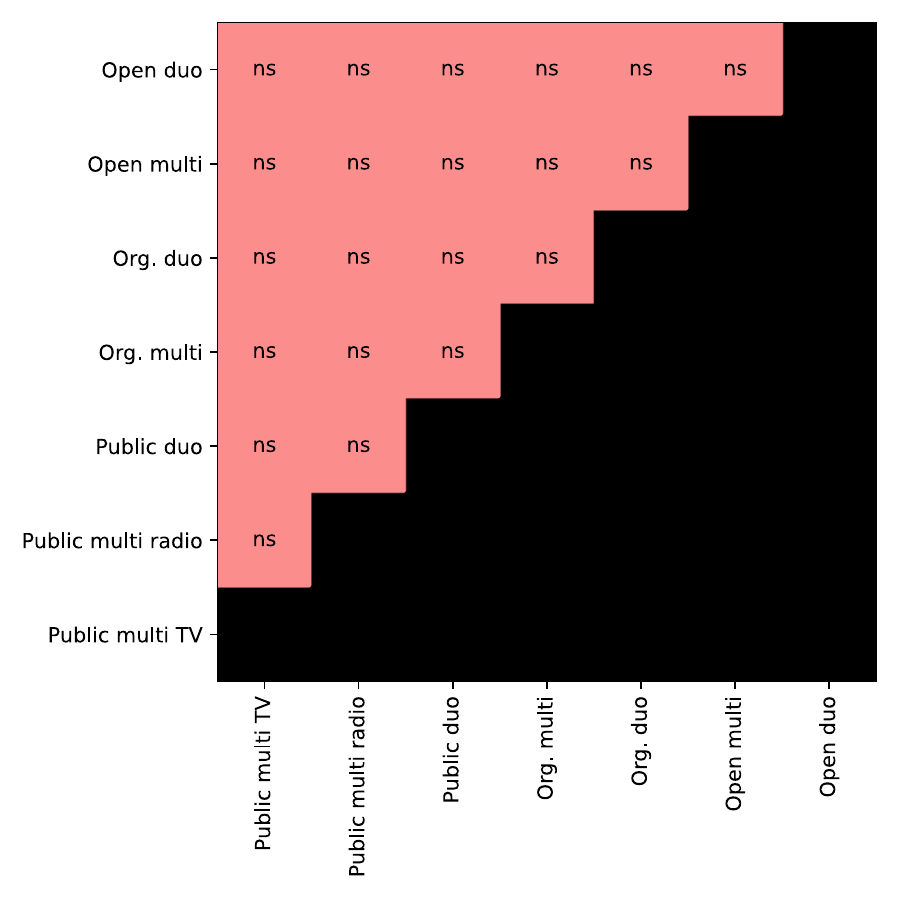}
            \caption{Gold annotations.}
            \label{figapp:salience_distribution_gold}
        \end{subfigure}
        \caption{Representation of the significance of the difference in salience distributions between dialogue group. Colours indicate the strength of the p-value.}
        \label{figapp:salience_distribution}
    \end{figure*}

    \paragraph{Salience.} \Cref{tabapp:avg_salience_answerability} reminds the average salience scores for each dialogue group, as well as the number of questions for which both annotators agreed on the score (support gold). \Cref{figapp:salience_distribution} shows the pairs of groups for which the differences in salience distributions are significant. For space reasons, we report the results for both annotators in \cref{figapp:salience_distribution_annotators} but each triangle should be analysed separately. 

    \begin{table*}[ht]
        \centering
        \begin{tabular}{l@{\hskip 3em}ccc@{\hskip 3em}ccc@{\hskip 3em}c}
            \toprule
            Group & S$_1$ & S$_2$ & S$_G$ & A$_1$ & A$_2$ & A$_G$ & Support Gold \\ 
            \midrule
            Open duo & 3.29 & 3.32 & 3.04 & .36 & .34 & .35 & 321/657 \\ 
            Open multi & 3.26 & 3.20 & 3.06 & .20 & .19 & .20 & 388/822 \\ 
            Org. duo & 4.04 & 4.12 & 4.08 & .47 & .40 & .44 & 742/1,548 \\ 
            Org. multi & 3.80 & 3.77 & 3.57 & .41 & .30 & .35 & 598/1,409 \\ 
            Public duo & 4.79 & 4.74 & 4.80 & .55 & .48 & .52 & 535/1,158 \\ 
            Public multi - radio & 4.22 & 3.88 & 3.95 & .56 & .44 & .50 & 402/768 \\ 
            Public multi - TV & 4.50 & 4.66 & 4.81 & .44 & .36 & .40 & 287/762 \\ 
            \midrule
            All & 3.92 & 3.90 & 3.82 & .42 & .35 & .38 & 3,273/7,124 \\ 
            \bottomrule
        \end{tabular}
        \caption{Average salience (0-8) and answerability (0-2) scores per dialogue group.}
        \label{tabapp:avg_salience_answerability}
    \end{table*}

    As explained in \cref{sec:salience}, most groups display significant differences in salience distributions. Open-domain duologues and multilogues are similar together while being different from all the other groups in both annotators' distributions. Public multilogues from the radio are similar to the organised duologues, while public multilogues from the television are similar to the public duologues, although the statistical strength of these differences varies by annotator. When looking into the differences for the gold annotations, we observe no significant differences between the groups of dialogues, which suggests that the disagreements between the annotators highlight more complex interactions, characteristic of the different dialogue types. This last result should be interpreted with caution, as the gold annotations represent a restricted subset of the data which limits the statistical power of the tests. Nevertheless, salience seems sensitive to dialogue type and in particular to the number of participants, with duologues displaying more salient questions than multilogues.


    \subsubsection{Answerability.} 
\begin{figure*}[hb!]
        \centering
        \begin{subfigure}{.85\columnwidth}
            \centering
            \includegraphics[width=\textwidth]{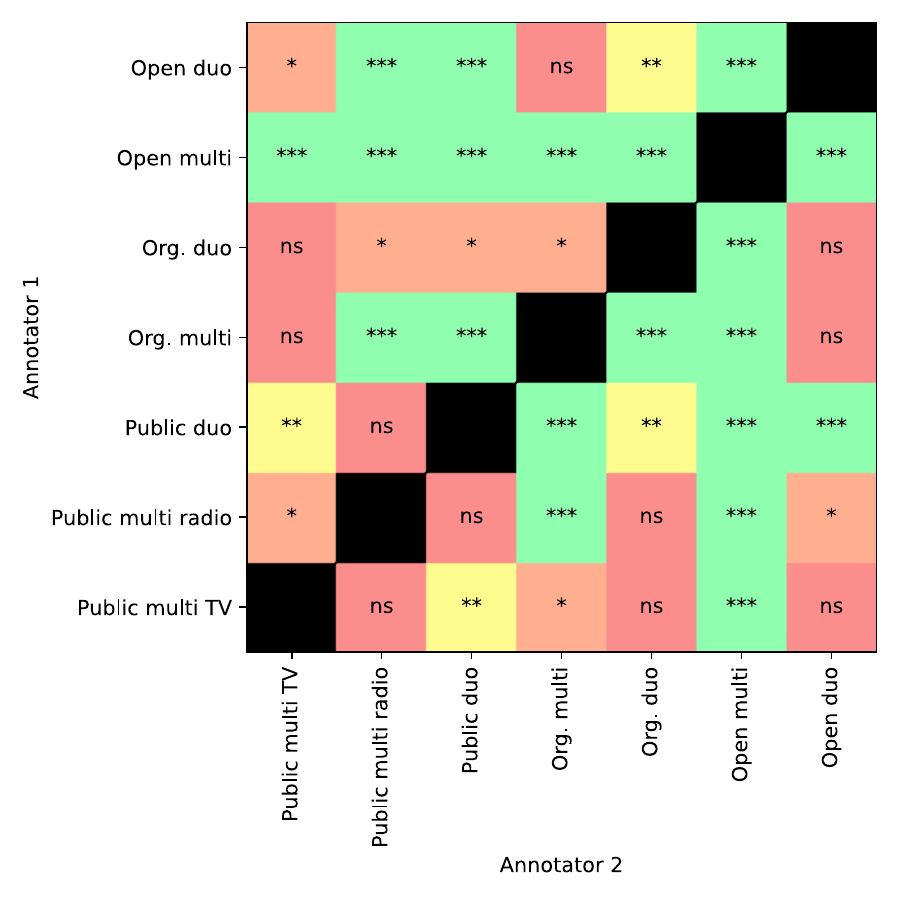}
            \caption{Annotators 1 (◤) and 2  (◢).}
            \label{figapp:answerability_distribution_annotators}
        \end{subfigure}
        \hfill
        \begin{subfigure}[b]{.3\columnwidth}
            \centering
            \includegraphics[width=\textwidth]{figures/legend_pvalues.pdf}
            \vspace{8em}
        \end{subfigure}
        \hfill
        \begin{subfigure}{.85\columnwidth}
            \centering
            \includegraphics[width=\textwidth]{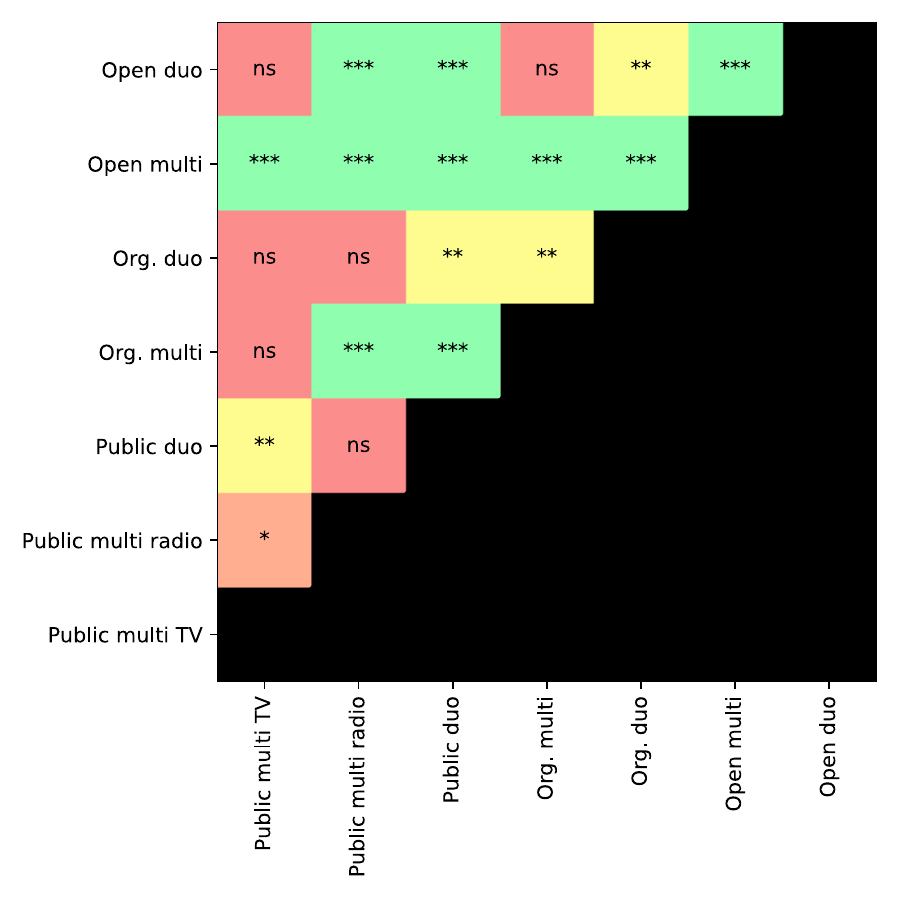}
            \caption{Gold annotations.}
            \label{figapp:answerability_distribution_gold}
        \end{subfigure}
        \caption{Representation of the significance of the difference in answerability distributions between dialogue group. Colours indicate the strength of the p-value.}
        \label{figapp:answerability_distribution}
    \end{figure*}

\Cref{tabapp:avg_salience_answerability} reminds the average answerability scores for each dialogue group. The gold scores are the average of the two annotators, hence we do not report the support contrary to salience. \Cref{figapp:answerability_distribution} shows the pairs of groups for which the differences in answerability distributions are significant. Again, for space reasons, we report the results for both annotators in \cref{figapp:answerability_distribution_annotators} but each triangle should be analysed separately. 
    
    As explained in \cref{sec:answerability}, answerability distributions are more stable than salience distributions across dialogue categories. The main variations come from open-domain multilogues, that are significantly different from all other dialogue types, and organised multilogues, that are different from all other types except open-domain duologues and public multilogues from television. Overall, these results suggest that answerability is less sensitive to dialogue type than salience. However, the low average answerability scores across \textit{all} groups indicates that most questions are not explicitly answered in the subsequent dialogue. 

    \subsection{Dialogue predictability}\label{secapp:predictability}

    \paragraph{Average answerability per salience score}

\begin{figure*}[ht]
        \centering
        \begin{subfigure}{.85\columnwidth}
            \centering
            \includegraphics[width=\columnwidth]{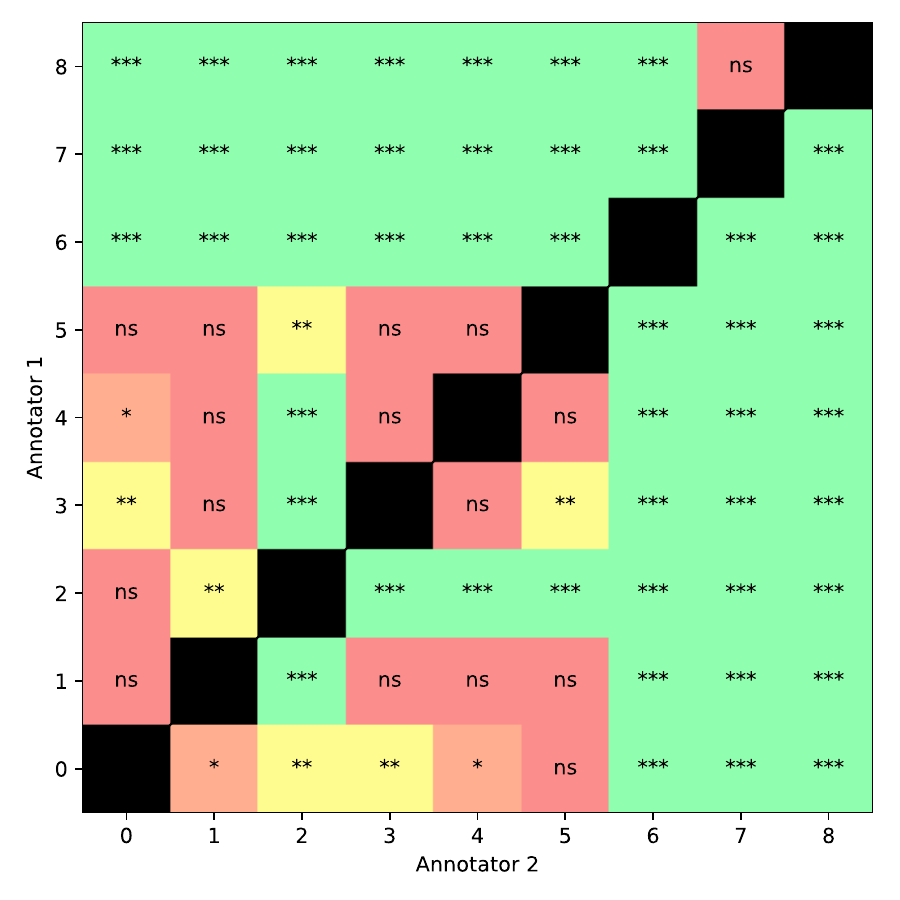}
            \caption{Annotators 1 (◤) and 2  (◢).}
            \label{figapp:avg_answerability_per_salience_annotators}
        \end{subfigure}
        \hfill
        \begin{subfigure}[b]{.3\columnwidth}
            \centering
            \includegraphics[width=\textwidth]{figures/legend_pvalues.pdf}
            \vspace{8em}
        \end{subfigure}
        \hfill
        \begin{subfigure}{.85\columnwidth}
            \centering
            \includegraphics[width=\columnwidth]{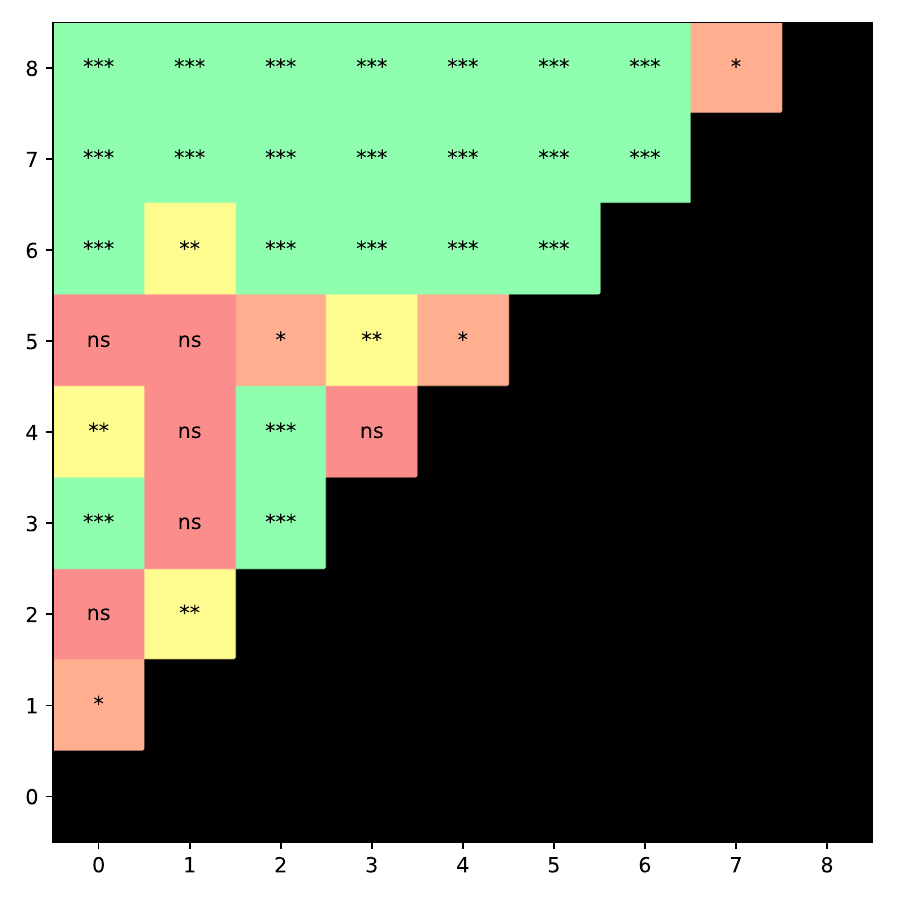}
            \caption{Gold annotations.}
            \label{figapp:avg_answerability_per_salience_gold}
        \end{subfigure}
        \caption{Representation of the significance of the difference in average answerability for each salience score. Colours indicate the strength of the p-value.}
        \label{figapp:avg_answerability_per_salience}
    \end{figure*}

    \Cref{tabapp:avg_answerability_per_salience} shows the average answerability score for each salience score and \cref{figapp:avg_answerability_per_salience} supports these results with a representation of the significance of these differences. The average answerability is relatively stable for salience scores below 6, with the exception of Score 2, which consistently shows lower answerability across conditions. Higher salience levels (6--8) are associated with significantly higher answerability, all different from each other except for Scores 7 and 8 with annotator 1. 

    \begin{table}
        \centering
        \begin{tabular}{cccc}
        \toprule
        S & A$_1$ & A$_2$ & A$_G$ \\
        \midrule
        0 & .23 $\pm$ .51 & .25 $\pm$ .55 & .18 $\pm$ .44 \\
        1 & .29 $\pm$ .60 & .33 $\pm$ .65 & .32 $\pm$ .64 \\
        2 & .17 $\pm$ .49 & .17 $\pm$ .49 & .13 $\pm$ .45 \\
        3 & .30 $\pm$ .60 & .33 $\pm$ .64 & .30 $\pm$ .60 \\
        4 & .30 $\pm$ .61 & .32 $\pm$ .63 & .33 $\pm$ .64 \\
        5 & .27 $\pm$ .56 & .28 $\pm$ .55 & .22 $\pm$ .50 \\
        6 & .52 $\pm$ .73 & .52 $\pm$ .74 & .53 $\pm$ .74 \\
        7 & 1.13 $\pm$ .81 & 1.11 $\pm$ .83 & 1.21 $\pm$ .81 \\
        8 & 1.03 $\pm$ .86 & .87 $\pm$ .85 & 1.41 $\pm$ .76 \\
        \bottomrule
        \end{tabular}
        \caption{Average answerability score per salience score. \textit{Results presented as mean $\pm$ standard deviation.}}
        \label{tabapp:avg_answerability_per_salience}
    \end{table}


            

        

        

    \paragraph{Correlation between salience and answerability}

    \begin{figure*}
        \centering
        \includegraphics[width=.7\textwidth]{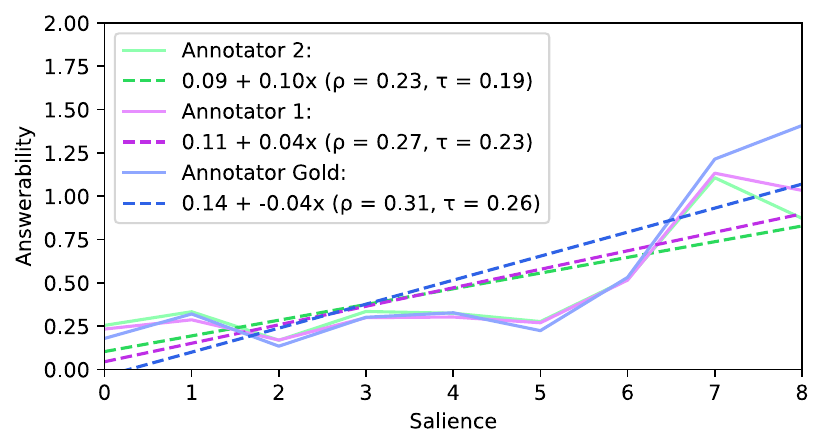}
        \caption{Correlation between salience and answerability scores ($\text{all p-values} \ll .001$) on the entire 0-8 scale. \textit{Legend in the same order as the dotted plots at 0.}}
        \label{figapp:correlation}
    \end{figure*}

    \Cref{figapp:correlation} reports the mean answerability per salience level along with regression lines, and the correlation coefficients in the legend for the full 0-8 scale. The coefficients are slightly higher than when the correlation is computed on the reduced scale [1, 3, 5, 6, 8] but the analysis provided in \cref{sec:correlation} remain valid. 

    \Cref{tabapp:avg_correlation} shows the correlation coefficient between salience and answerability for each dialogue group. As explained in \cref{sec:correlation} when ordering the groups by their $\rho$ coefficient, we observe that both annotators have a very similar ranking, with only two groups swapping their position: the open-domain duologues and the public multilogues from the television that are ranked second and fifth based on annotator 1 in increasing order, and fifth and second based on annotator 2. This suggests that the differences between groups are relatively consistent across annotators but that these specific two groups may be more or less complex to annotate depending on the annotator. 
    However, this ranking is very different from the one obtained on the gold annotations, where only the two groups with highest $\rho$ coefficients -- the organised duologues and the public duologues -- keep the same ranking. 

\begin{figure}[h!]
        \centering
        \begin{subfigure}{.8\columnwidth}
            \centering
            \caption{Annotators 1 (◤) and 2  (◢).}

            \vspace{.1em}
            
            \includegraphics[width=\columnwidth]{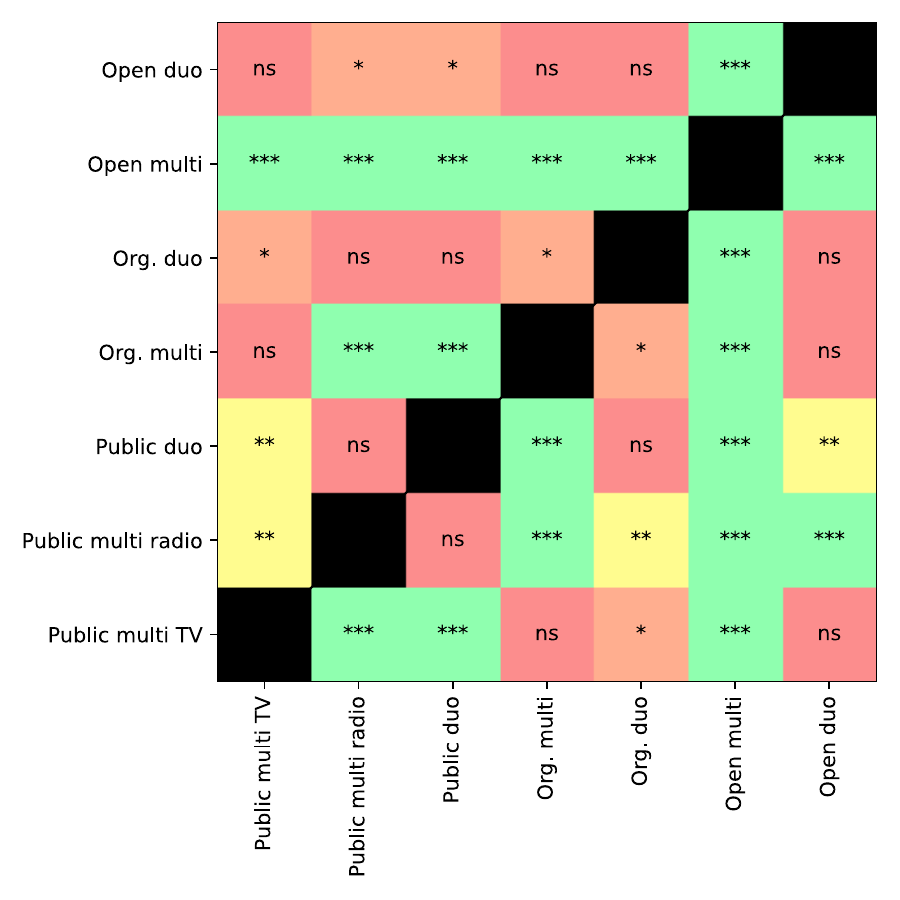}
            \label{figapp:correlation_diff_groups_annotators}
        \end{subfigure}

        \vspace{-1.2em}
        
        \begin{subfigure}[b]{.8\columnwidth}
            \centering
            \includegraphics[width=.96\columnwidth]{figures/legend_pvalues_horiz.pdf}
        \end{subfigure}

        \vspace{.2em}
        
        \begin{subfigure}{.8\columnwidth}
            \centering
            \includegraphics[width=\columnwidth]{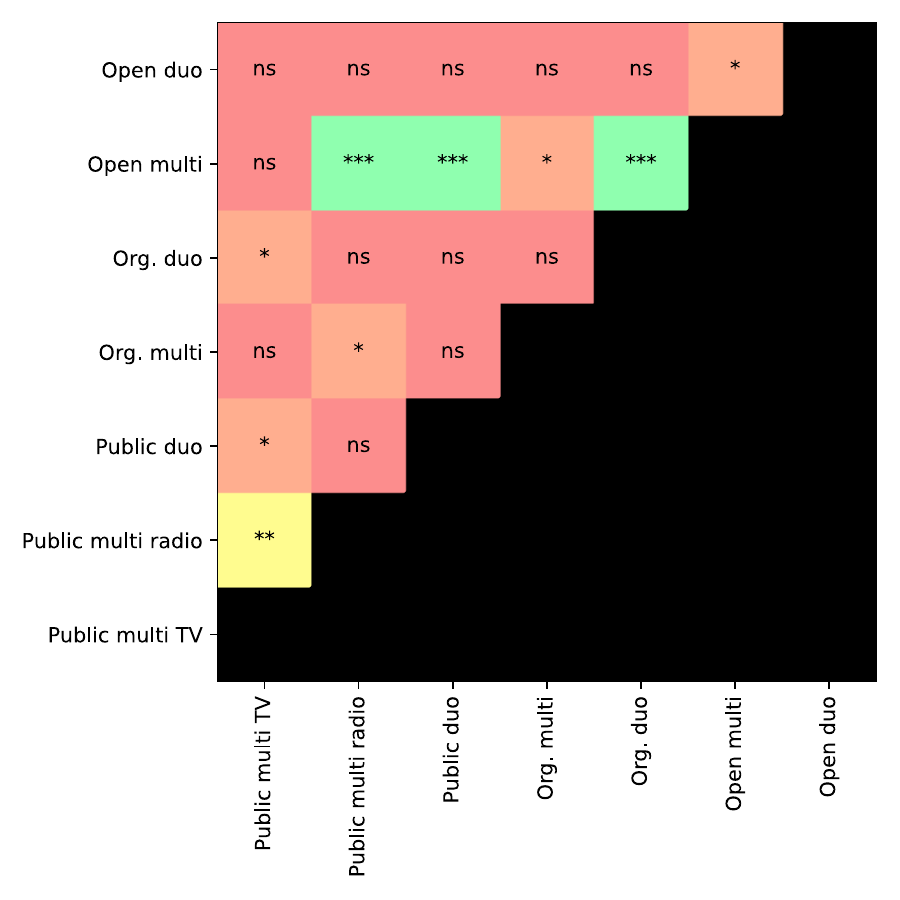}
            \caption{Gold annotations.}
            \label{figapp:correlation_diff_groups_gold}
        \end{subfigure}
        \caption{Representation of the significance of the difference in correlation between dialogue group. Colours indicate the strength of the p-value.}
        \label{figapp:correlation_diff_groups}
    \end{figure}

    Nevertheless, \cref{figapp:correlation_diff_groups} shows that most group differences in correlation are not statistically significant with the exception of open-domain multilogues that have the lowest $\rho$ coefficient. Public multilogues from the television, with the intermediate $\rho$ coefficient for both annotators, also show significant differences with most other groups, open-domain duologues and organised multilogues being the only groups similar to them. Moreover, there are even less differences between groups when looking at the gold annotations. The homogenisation of the results in the gold annotations suggests that the disagreements between the annotators reveal some meaningful differences in the dialogues. 

    \begin{table}
        \centering
        \begin{tabular}{lccc}
            \toprule
            Group & $\rho_1$ & $\rho_2$ & $\rho_G$ \\
            \midrule
            ALL & 0.244 & 0.193 & 0.239 \\
            Open duo & \textit{0.148} & (0.107) & (0.129) \\
            Open multi & \textit{0.168} & 0.262 & \textit{0.260} \\
            Org. duo & 0.226 & 0.217 & 0.248 \\
            Org. multi & 0.232 & 0.198 & 0.225 \\
            Public duo & 0.315 & 0.215 & 0.323 \\
            Public multi - radio & 0.216 & \textit{0.114} & (0.098) \\
            Public multi - TV & 0.265 & 0.163 & 0.306 \\
            \bottomrule
        \end{tabular}
        \caption{Correlation between salience and answerability by dialogue group, computed on questions whose salience score was 1, 2, 4, 6, or 8. $\rho_i$ refers to the Spearman's rank correlation coefficient and associated p-value for annotator $i$, while $\rho_G$ refers to the same metrics for the gold annotations. Most correlations had an associated p-value below 0.001, indicating statistical significance. The numbers in (parenthesis) indicate non-significant results with a p-value above 0.05, and \textit{italics} indicates a weak p-value between 0.001 and 0.05.}
        \label{tabapp:avg_correlation}
    \end{table}

\end{document}